\documentclass[journal,11pt]{IEEEtran}
\usepackage{cite}
\usepackage{url}  
\usepackage[english]{babel}
\usepackage{CJK}
\usepackage{multirow}
\usepackage{graphicx}  
\ifCLASSINFOpdf
\else
\fi
\begin{document}
%
\title{Deep Stacking Networks for Low-Resource Chinese Word Segmentation with Transfer Learning}
%
%
%

\author{Jingjing Xu, Xu Sun, Sujian Li, Xiaoyan Cai and Bingzhen Wei
}

\maketitle

\begin{abstract}
In recent years, neural networks have proven to be effective in Chinese word segmentation. However, this promising performance relies on large-scale training data. Neural networks with conventional architectures cannot achieve the desired results in low-resource datasets due to the lack of labelled training data. In this paper, we propose a deep stacking framework to improve the performance on word segmentation tasks with insufficient data by integrating datasets from diverse domains. Our framework consists of two parts, domain-based models and deep stacking networks. The domain-based models are used to learn knowledge from different datasets. The deep stacking networks are designed to integrate domain-based models. To reduce model conflicts, we innovatively add communication paths among models and design various structures of deep stacking networks, including Gaussian-based Stacking Networks, Concatenate-based Stacking Networks, Sequence-based Stacking Networks and Tree-based Stacking Networks. We conduct experiments on six low-resource datasets from various domains. Our proposed framework shows significant performance improvements on all datasets compared with several strong baselines.
\end{abstract}

\begin{IEEEkeywords}
Deep Stacking Networks, Chinese Word Segmentation, Transfer Learning.
\end{IEEEkeywords}

\section{Introduction}

Chinese word segmentation (CWS) is an important step in Chinese language processing, which is usually treated as a sequence labelling problem. Formally, given an input sequence of Chinese characters $\mathbf{x}=x_{1}x_{2}...x_{n}$, the goal is to produce a sequence of tags $\mathbf{y}=y_{1}y_{2}...y_{n}$. Many traditional statistic methods have achieved high accuracies on the news domain~\cite{Lafferty,Tseng05,ZhaoHLL10,SunZMTT09,SunWL12,SunZMTT13,ZhangWSM13,SunLWL14,Sun_NIPS2014}. However, these approaches incorporate many handcrafted features, making it hard to generalize broadly.

Recently, deep neural networks have made significant improvements in Chinese word segmentation. These works aim at automatically extracting features. Collobert et
al. \cite{collobert} developed a general neural network architecture for sequence labelling tasks. Following
this work, neural network approaches have been
well studied and widely applied to CWS with
good results ~\cite{pei,chen,Zhang2016Transition}.
 Pei et al.~\cite{pei} used Convolutional Neural Network (CNN) to capture local information within a fixed size window and proposed a tensor framework to capture the information of previous tags. Chen et al.~\cite{chen} proposed Gated Recursive Neural Network (GRNN) to model feature combinations of context characters. Zhang et al.~\cite{Zhang2016Transition} proposed a neural model for word-based CWS by replacing the manually designed features with neural features
in a word-based segmentation framework.

\begin{table}
\caption{ Results of Bi-LSTM on high-resource and low-resource datasets. For low-resource datasets, there is a clear result decline compared with high-resource datasets, MSR and CTB8. }
\centering

\begin{tabular}{c|c|c|c|c}
\hline

\multicolumn{1}{c}{Datasets}&\multicolumn{1}{c}{Train}&\multicolumn{1}{c}{Test}&\multicolumn{1}{c}{F-score}\\
\hline
\multicolumn{1}{c}{\multirow{1}{*}{MSR}}&\multicolumn{1}{c}{\multirow{1}{*}{86918}}&\multicolumn{1}{c}{\multirow{1}{*}{3985}}& \multicolumn{1}{c}{95.04}\\

\multicolumn{1}{c}{\multirow{1}{*}{CTB8}}&\multicolumn{1}{c}{\multirow{1}{*}{68918}}&\multicolumn{1}{c}{\multirow{1}{*}{2193}}& \multicolumn{1}{c}{94.38}\\
\hline
\hline
\multicolumn{1}{c}{\multirow{1}{*}{Social Media}}&\multicolumn{1}{c}{\multirow{1}{*}{20135}}&\multicolumn{1}{c}{\multirow{1}{*}{8592}}&\multicolumn{1}{c}{91.00}\\


\multicolumn{1}{c}{\multirow{1}{*}{Disease}}&\multicolumn{1}{c}{\multirow{1}{*}{5321}}&\multicolumn{1}{c}{\multirow{1}{*}{1330}}&\multicolumn{1}{c}{87.66}\\

\multicolumn{1}{c}{\multirow{1}{*}{Tourism}}&\multicolumn{1}{c}{\multirow{1}{*}{4496}}&\multicolumn{1}{c}{\multirow{1}{*}{1124}}&\multicolumn{1}{c}{91.36}\\

\multicolumn{1}{c}{\multirow{1}{*}{Patent}}&\multicolumn{1}{c}{\multirow{1}{*}{916}}&\multicolumn{1}{c}{\multirow{1}{*}{228}}&\multicolumn{1}{c}{87.39}\\

\multicolumn{1}{c}{\multirow{1}{*}{Fiction}}&\multicolumn{1}{c}{\multirow{1}{*}{534}}&\multicolumn{1}{c}{\multirow{1}{*}{113}}&\multicolumn{1}{c}{85.48}\\

\hline
\end{tabular}

\label{resultintroduction}
\end{table}

\begin{figure}
  \centerline{\includegraphics[width=\linewidth]{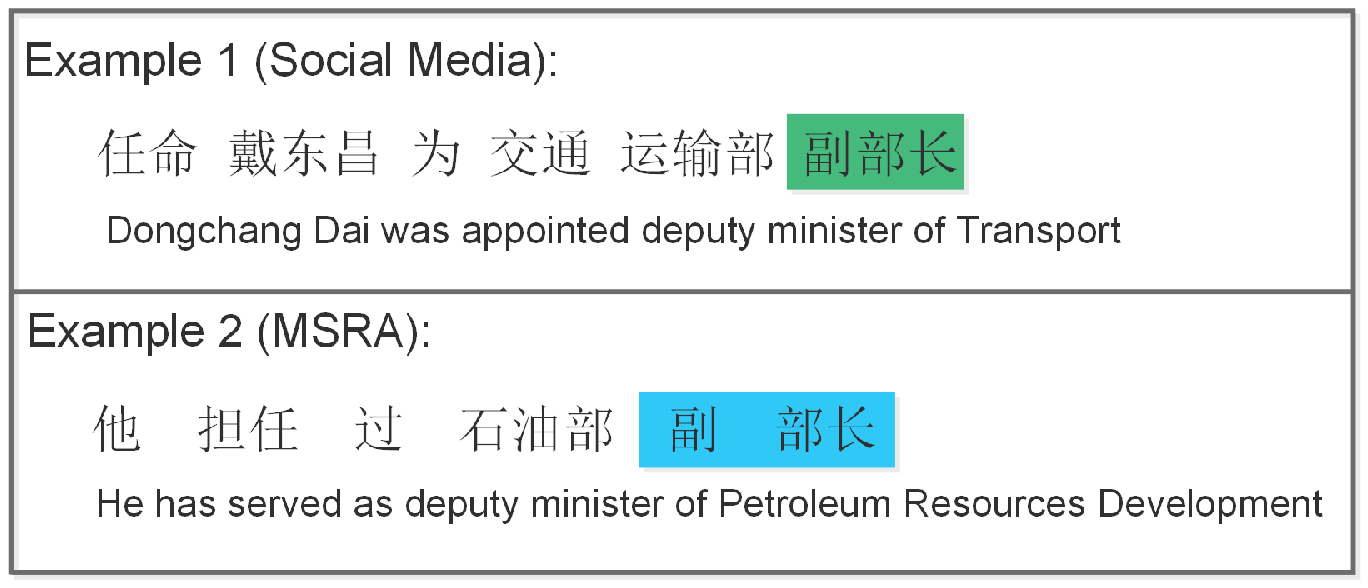}}
  \caption{An example of incompatible annotation guidelines. The MSRA dataset treats deputy minister as one word (shown in blue) and the Social Media dataset treats deputy minister as two words (shown in green). }
\label{problem1}
\end{figure}

All of the models above achieve very promising results, however, their success relies on large-scale annotated datasets. Neural networks with conventional architectures fail in low-resource datasets due to the lack of labelled data. In this paper, we choose a widely used neural network, Bi-directional Long Short Term Memory Network (Bi-LSTM), to show this failure. We conduct experiments on two high-resource datasets and several low-resource datasets. The datasets will be introduced in the experiment section. Table \ref{resultintroduction} shows that there is a large performance gap between high-resource and low-resource datasets. It proves that insufficient annotated data poses obstacles for training a powerful neural word segmentation model. Fortunately, there are a number of multi-domain corpora which can be used to improve the performance. Thus, in this paper, we propose to incorporate multi-domain knowledge to deal with the lack of training data. However, simply borrowing knowledge from other domains probably results in ``Dataset Bias".  First, a model loses accuracy when the target data distribution is different with the source data distribution. Second, for some words, different corpora have incompatible annotation guidelines as shown in Figure \ref{problem1}. ``Dataset Bias'' leads to model conflicts  which make it difficult to directly integrate models.

To address these problems, we propose a novel framework which consists of two parts, domain-based models and deep stacking networks (DSNs). The domain-based models are pre-trained on different datasets to make use of the large scale data to learn segmentation knowledge. The deep stacking networks are designed to integrate domain-based models. To reduce model conflicts, communication among models is necessary. Thus, we add communication paths to increase the interaction among models and design various stacking networks, including Gaussian-based Stacking Networks (Gaussian-SNs), Concatenate-based Stacking Networks (Concatenate-SNs), Sequence-based Stacking Networks (Sequence-SNs) and Tree-based Stacking Networks (Tree-SNs).  Gaussian-SNs, Concatenate-SNs and Sequence-SNs belong to the same category. The intuition of this category is to dynamically decide which model is more suitable for the case and should be assigned with a higher weight explicitly. Different from these three networks, Tree-SNs use a recursive structure as a replacement of weighted voting to integrate models implicitly.  


We conduct experiments on six low-resource datasets from different domains. Experimental
results demonstrate that our methods outperform several strong baselines in all datasets. We also find that Sequence-SNs are suitable to most of datasets and Tree-SNs are suitable to datasets which have extremely insufficient training data.  

The major contributions of this paper are listed as follows:
\begin{itemize}

\item We propose a novel stacking framework to improve low-resource CWS by integrating multi-domain knowledge.

\item To decrease model conflicts, we add communication paths among models and design various deep stacking networks.

\item Experiment results show that our methods significantly outperform competitive baselines in all datasets. Furthermore, we  find that Sequence-SNs are suitable to most of datasets and Tree-SNs are suitable to datasets which have extremely insufficient training data.

%
%
\end{itemize}

 \begin{figure}
  \centerline{\includegraphics[width=\linewidth]{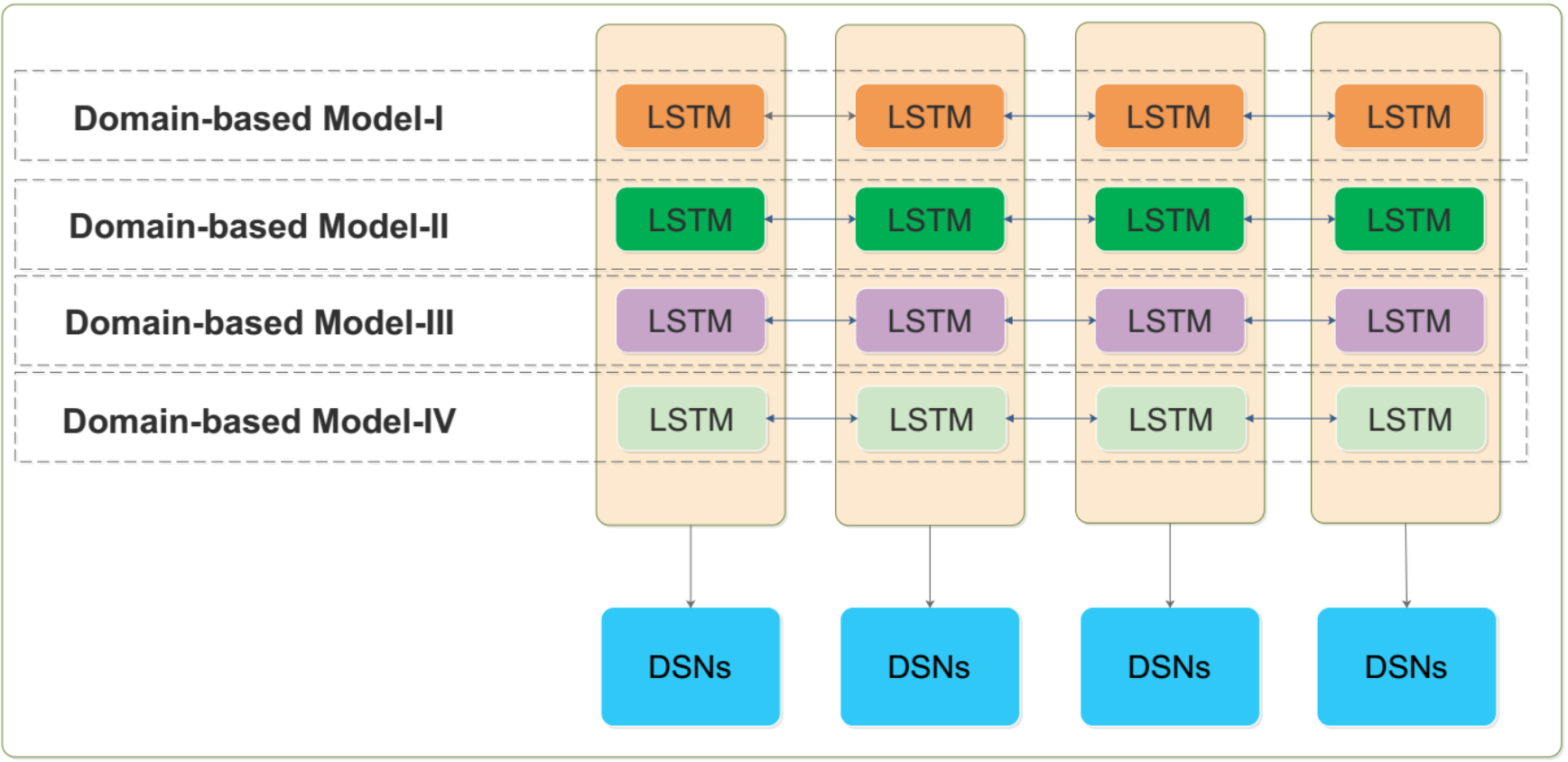}}
\caption{An illustration of the proposed framework. The whole framework consists of two parts, domain-based models and deep stacking networks (DSNs).} 
\label{framework}
\end{figure}

\section{Deep Stacking Framework}



\begin{figure*}
  \centerline{\includegraphics[width=0.8\textwidth  ]{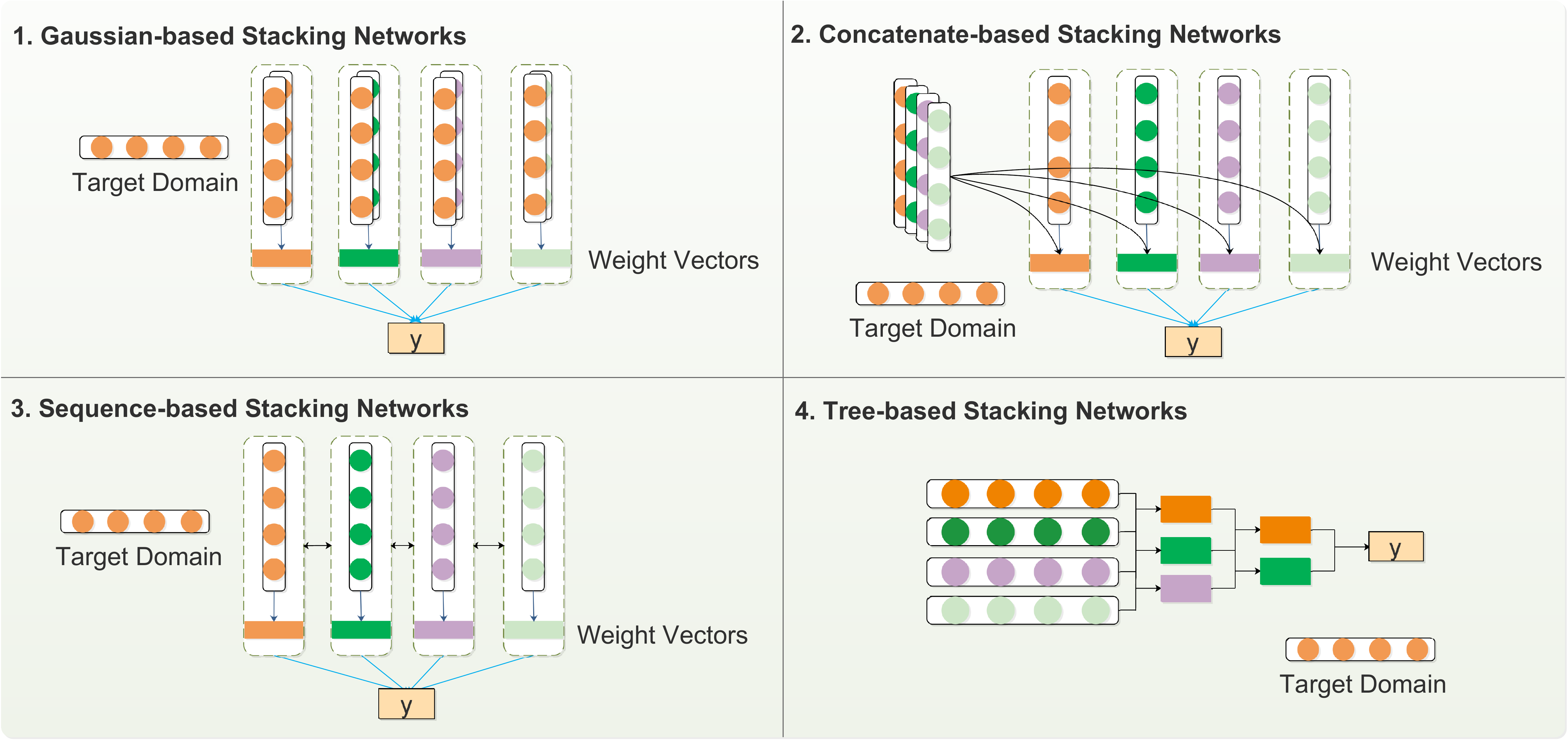}}
\caption{An illustration of four DSNs. The orange vector represents the output of the model which is pre-trained on a target domain. Gaussian-SNs, Concatenate-SNs and Sequence-SNs encode weight vectors explicitly and produce the output by the way of weighted voting. Gaussian-SNs produce weight vectors based on the similarities between each domain and a target domain. Concatenate-SNs produce weight vector based on the concatenation of the outputs of models. Sequence-SNs produce weight vectors in a sequence structure.  Tree-SNs adopt a recursive structure to combine the outputs of domain-based models, rather than weighted voting.   }
\label{dsns}
\end{figure*}

In this paper, we propose a deep stacking framework (shown in Figure \ref{framework}). The whole framework consists of two parts, domain-based models and deep stacking networks. The domain-based models are first trained on different datasets, respectively. The deep stacking networks focus on incorporating these heterogeneous models. 

\subsection{Domain-based Models} To avoid the shift of distributions and the conflicts of annotation guidelines, we do not involve out-of-domain corpora in the training process directly. Instead, we separately pre-train each domain-based model. Furthermore, we do not limit domain-based models to any specific structure. It can be applied in any sequence labelling model, such as RNN, Bi-LSTM and so on. 

\subsection{Deep Stacking Networks}

To reduce model conflicts, we add communication paths among models and propose various deep stacking networks, as shown in Figure \ref{dsns}. Gaussian-SNs, Concatenate-SNs and Sequence-SNs can be categorized into the same group. This group encodes weight vectors explicitly and produces the output by the way of weighted voting. Unlike bagging  which has fixed voting strengths, this group dynamically computes weight vectors given different inputs. The major difference among the three networks lies in the way of producing weight vectors. Gaussian-SNs produce weight vectors based on the similarities between each domain and a target domain. Concatenate-SNs produce weight vector based on the concatenation of the outputs of models. Sequence-SNs produce weight vectors in a sequence structure. Unlike the models mentioned above, Tree-SNs adopt a recursive structure to combine the outputs of domain-based models, rather than weighted voting.

Assume we are provided with several domain-based models, called $S_{1}, S_{2}, ......, S_{m}$ where $m$ is the number of models. For a word $x_{i}$, the inputs to DSNs are hidden states of domain-based models, $h_{i,1}, ......, h_{i,m}$. The target of DSNs is to integrate domain-based models and produce the output $y$, like meta-learning. 

%

%





\textbf{Gaussian-based Stacking Networks (Gaussian-SNs)}
The core idea is that the weights are decided by the similarities between each domain and a target domain. The Gaussian kernel function is a very interesting and powerful tool to calculate similarities between two vectors. In this paper, we use this function to calculate domain similarities as
\begin{equation}
S(h_{i,j},h_{i,t})=\frac{1}{Z}exp(-\frac{||h_{i,j}-h_{i,t}||}{2\sigma^{2}})
\end{equation}
where $h_{i,j}$ and $h_{i,t}$ are the outputs of the models produced by a domain $x$ and  a target domain $t$, respectively. $\sigma$ is used to control the variance of the Gaussian kernel function. $Z$ is a regularization term and calculated as
\begin{equation}
Z=\sum_{j=1}^{m}exp(-\frac{||h_{i,j}-h_{i,t}||}{2\sigma^{2}})
\end{equation}
where $Z$ is designed to make the sum of entire probability distributions equal to $1$.

After producing the weight vectors, the output $y_{i}$ is calculated as
\begin{equation}
y_{i}=softmax(\sum_{j=1}^{m} S(h_{i,j},h_{i,t}) \cdot h_{i,j})
\end{equation}
where $h_{i,j}$ is the output of $j_{th}$ model and $S(h_{i,j},h_{i,t})$ is the weight of $h_{i,j}$.

The major drawback of this non-parametric model is that the performance is mainly affected by $\sigma$ which is set manually. On the one hand, we need to try a number of values on development datasets to choose the best one. On the other hand, the best value is not applicable to every dataset and we have to adjust the parameter repeatedly.

In order to solve this problem, we propose a parametric model, called Concatenate-SNs, where all parameters are self-learned on the process of training.  


\textbf{Concatenate-based Stacking Networks (Concatenate-SNs)} The concatenation among all outputs of models is taken into consideration when weight vectors are being computed. The intuition is to first mix all information together and then calculate weight vectors given different inputs. The detailed calculating process is described as follows. 

For word $w_{i}$, the output $y_{i}$ is computed as
\begin{equation}
y_{i}=softmax(g(s_{i}))
\end{equation} 
where $g$ is a non-linear function, $s_{i}$ is calculated based on weighted sum of hidden-layer outputs of models as
\begin{equation}
s_{i}=\sum_{j=1}^{m}{\alpha}_{i,j}\cdot h_{i,j}
\end{equation} 
where $h_{i,j}$ is the output vector of the $j_{th}$ model. The weighted vector ${\alpha}_{i,j}{\in}R^{d}$ is computed as
\begin{equation}
\alpha_{i,j}=\frac{exp(e_{i,j})}{\sum_{p=1}^{m}exp(e_{i,p})}
\end{equation}
where $e_{i,j}{\in}R^{d}$ is a vector and computed as
\begin{equation}
e_{i,j}=f(W_{1}[W_{2} \cdot \mathbf{h}_{i},h_{i,j}])
\end{equation}
where $f$ is a non-linear function, $W_{1}{\in}R^{d \times 2d}$ and $W_{2}{\in}R^{d \times md}$ are parameter metrics to be learned, $d$ is the dimension of hidden states, $[...,...]$ denotes vector concatenation, $\mathbf{h}_{i}=\{h_{i,1},...h_{i,j},...,h_{i,m}\}$ is the concatenation of hidden states. All parameter metrics are trained on a target dataset. We refer to this network as \textbf{Concatenate-SNs}. 



It makes sense to take all necessary information into consideration when weight vectors are being produced. However, simply concatenating all information into a single vector probably loses key information or introduces new noise information. Next, we design a new model, called Sequence-SNs, by introducing a new mechanism to distinguish key information from noise vectors automatically.

\textbf{Sequence-based Stacking Networks (Sequence-SNs)} 
In this networks, weight vectors are produced in a sequence mode.  First, the output of each domain-based model $h_{i,j}$  is processed by Bi-LSTM which produces a weight vector $\alpha_{i,j}$. Bi-LSTM is used to decide which information should be kept. The intuition is to guarantee that only the key information are transmitted in recurrent cells. Finally, the output $y_{i}$ is generated by weighted voting. The detailed calculating process is described as follows. 

The output $y_{i}$ is calculated as
\begin{equation}
y_{i}=softmax(\sum_{j=1}^{m} \alpha_{i,j}\cdot h_{i,j})
\end{equation}
where $h_{i,j}$ is the output vector of the $j_{th}$ model. The weighted vector ${\alpha}_{i,j}$ is computed as
\begin{equation}
\alpha_{i,j}=\frac{exp(e_{i,j})}{\sum_{p=1}^{m}exp(e_{i,p})}
\end{equation}
where $e_{i,j}$ is the output of Bi-LSTM.

Bi-LSTM produces the tag sequences by generating one target tag $y_{i}$ at each time step via a softmax layer. The next output depends on the previous hidden state. Three different gates are computed according to the state and output of previous cell, $s^{(t-1)}$ and $h^{(t-1)}$.
\begin{eqnarray}
g^{(t)}=\tanh(w_{gx} x^{(t)}+w_{gh} h^{(t-1)}+b_{g})
\end{eqnarray}
\begin{eqnarray}
i^{(t)}=sigmoid(w_{ix} x^{(t)}+w_{ih} h^{(t-1)}+b_{i})
\end{eqnarray}
\begin{eqnarray}
f^{(t)}=sigmoid(w_{fx} x^{(t)}+w_{fh} h^{(t-1)}+b_f )
\end{eqnarray}
\begin{eqnarray}
o^{(t)}=sigmoid(w_{ox} x^{(t)}+w_{oh} h^{(t-1)}+b_o)
\end{eqnarray}

The core of LSTM cell is $s^{(t)}$, which is computed by the former state $s^{(t-1)}$ and two gates, ${i^{(t)}}$ and $f^{(t)}$.
\begin{eqnarray}
s^{(t)}=g^{(t)}\odot i^{(t)}+s^{(t-1)}\odot f^{(t)}
\end{eqnarray}

Finally, the output of LSTM cell $e_{i,j}$ is calculated by $s$ and $o$.
\begin{eqnarray}
e_{i,j}=\tanh(s\odot o)
\end{eqnarray}

Considering the order of models probably affects the performance, we randomly shuffle the order and report the result variance in the experiments section. 

All networks mentioned above are under the framework of weighted voting. However, we are still curious about if the model can learn how to integrate models autonomously. Thus, we propose Tree-SNs by introducing a new structure as a replacement of weighted voting.

\textbf{Tree-based Stacking Networks (Tree-SNs)} This networks drop the framework of weighted voting and combine the outputs of domain-based models recursively. The core idea is to learn how to make decisions when the two inputs are conflicting. Recursive cells share the same parameters in the same layer.

In each cell of recursive tree, output $h_{j}^{l}{\in}R^{d}$ of the $j_{th}$ hidden node at recursive layer $l$ is computed as
\begin{eqnarray}
h_{j}^{l}=z_{N}{\odot}h^{'}+z_{L}{\odot}h^{l-1}_{j}+z_{R}{\odot}h^{l-1}_{j+1}
\end{eqnarray}
where $z_{N}$, $z_{L}$ and $z_{R}$ are update gates. ${\odot}$ means element-wise multiplication. To simplify the cell of recursive tree, $z_{N}$, $z_{L}$ and $z_{R}$ are computed as
\begin{eqnarray}
\left[
\begin{array}{c}
z_{N}  \\
z_{L} \\
z_{R}
\end{array}
\right]
=sigmoid(U
\left[
\begin{array}{c}
 h^{'}\\
 h^{l-1}_{j}\\
 h^{l-1}_{j+1}
\end{array}
\right]
 )
\end{eqnarray}
where $U{\in}R^{3d \times 3d}$ is the coefficient of the update gates. $h^{'}$ is computed as
\begin{eqnarray}
h^{'}
=tanh(W
\left[
\begin{array}{c}
 r_{L}{\odot}h^{l-1}_{j}\\
 r_{R}{\odot}h^{l-1}_{j+1}
\end{array}
\right]
 )
\end{eqnarray}
where $W{\in}R^{d \times 2d}$ is the coefficient of the reset gates. $r_{L}{\in}R^{d \times d}$ and $r_{R}{\in}R^{d \times d}$ are the reset gates for nodes $h^{l-1}_{j}$ and $h^{l-1}_{j+1}$, which can be formalized as
\begin{eqnarray}
\left[
\begin{array}{c}
r_{L} \\
r_{R}
\end{array}
\right]
=sigmoid(G
\left[
\begin{array}{c}
 h^{l-1}_{j}\\
 h^{l-1}_{j+1}
\end{array}
\right]
 )
\end{eqnarray}
where reset gates control the selection of left and right children to produce the current new activation $h^{'}$. The activation of an output neuron can be regarded as a choice among $h^{'}$,  $h^{l-1}_{j}$ and $h^{l-1}_{j+1}$.

\section{Training}
\label {al}

%
%
The objective is to minimize the negative log-likelihood. The loss function is 
\begin{eqnarray}
J(\theta)=-\frac{1}{N}\sum_{i=1}^{N}\sum_{j=1}^{m}{log(p(y_{i,j}|X_{i},\theta))}
\end{eqnarray}
where $p(y_{i,j}|X_{i},\theta)$ is the conditional probability of output word $y_{i,j}$ given source texts $X_{i}$.

We find that Adam~\cite{Kingma2014Adam} is a practical method to train large neural networks. Then, we conduct experiments on Adam training algorithm.

First, it recursively calculates $m_{t}$ and $v_{t}$, based on the gradient $g_{t}$. $\beta_{1}$ and $\beta_{2}$ aim to control the ratio of previous states.

\begin{eqnarray}
m_{t}=\beta_{1}\cdot m_{t-1}+(1-\beta_{1})\cdot g_{t}
\end{eqnarray}
\begin{eqnarray}
v_{t}=\beta_{2}\cdot v_{t-1}+(1-\beta_{2})\cdot g_{t}^{2}
\end{eqnarray}
Second, it calculates $\Delta W(t)$ based on $v_{t}$ and $m_{t}$. $\epsilon$ and $\mu$ both are smooth parameters.

\begin{eqnarray}
M(w,t) =v_{t}-m_{t}^{2}
\end{eqnarray}
\begin{eqnarray}
\Delta W(t)=\frac{\epsilon g_{t,i}}{\sqrt{M(w,t)}+\mu}
\end{eqnarray}
Finally, the parameter update for the $\Theta_{t,i}$ at time step $t$ on $i_{th}$ layer is

\begin{eqnarray}
\Theta_{t,i}=\Theta_{t,i}-\Delta W(t)
\end{eqnarray}
Following experiment results on a development dataset, hyper parameters of optimization method are set as follows: $\beta_{1}=\beta_{2}=0.95$, $\epsilon=1\times10^{-2}$.

\begin{table}
\caption{Details of datasets.
}
\centering
\begin{tabular}{|c|c|c|c|c|}

\hline
\multicolumn{2}{|c|}{\multirow{1}{*}{Datasets}}&Words&Chars&Sents\\

\hline

\multicolumn{1}{|c|}{\multirow{2}{*}{MSRA}}&{Train}&2368391&3990584&86918\\
&Test&106873&181487&3985\\
\hline
\multicolumn{1}{|c|}{\multirow{2}{*}{CTB8}}&{Train}&1553943&2445304&68918\\
&Test&60142&99159&2193\\

\hline
\multicolumn{1}{|c|}{\multirow{2}{*}{Social Media}}&{Train}&421155&657605&20135\\
&Test&187856&302014&8592\\
\hline
\multicolumn{1}{|c|}{\multirow{2}{*}{PKU}}&{Train}&1109947&1794871&19054\\
&Test&104372&169512&1944\\
\hline

\multicolumn{1}{|c|}{\multirow{2}{*}{Disease}}&{Train}&87709&155777&5321\\
&Test&21109&37784&1330\\
\hline

\multicolumn{1}{|c|}{\multirow{2}{*}{Tourism}}&{Train}&29655&41805&4496\\
&Test&7569&10790&1124\\

\hline
\multicolumn{1}{|c|}{\multirow{2}{*}{Patent}}&{Train}&28818&45029&916\\
&Test&7903&12217&228\\

\hline

\multicolumn{1}{|c|}{\multirow{2}{*}{Fiction}}&{Train}&15914&22777&534\\
&Test&4470&6476&133\\

\hline
\end{tabular}

\label{datasets}
\end{table}

\section{Experiments}

To demonstrate the effectiveness of our proposed methods, we conduct experiments on six low-resource datasets. In this section we will describe the details of datasets, experiment settings, and the baseline methods we compare with.

\subsection{Datasets}

Table \ref{datasets} gives the details of eight datasets which are used in experiments. They are from different domains, including social media, news, tourism, disease, fiction and patent. We use 10\% data of shuffled training set as the development
set for all datasets. 
\begin{itemize}

\item \textbf{PKU and MSRA}. PKU and MSRA are provided by the second International Chinese Word Segmentation Bakeoff (Emerson, 2005). The two datasets are from traditional news domain. 

\item \textbf{CTB8} Chinese Treebank 8.0\footnote{https://catalog.ldc.upenn.edu/ldc2013t21} consists of approximately 1.5 million words of annotated and parsed texts from Chinese news, government documents, magazine articles, various broadcast news and broadcast conversation programs, web newsgroups and weblogs. 

\item \textbf{Disease, Fiction, Tourism and Patent}  The corpus is originally constructed in Qiu et al.~\cite{LiqunQiu} by annotating multi-domain texts. We build four low-resource datasets based on this corpus. Each of these datasets has extremely insufficient training data. 

\item \textbf{Social Media} Social media texts are from Sina Weibo\footnote{Sina Weibo is a Chinese microblogging (weibo) website.}, which are provided by NLPCC 2016 shared task\footnote{http://tcci.ccf.org.cn/conference/2016/}. This task aims to evaluate the techniques of Chinese word segmentation for Weibo texts. 

\end{itemize}

As shown in Table \ref{datasets}, we use MSRA and CTB8 as high-resource datasets and the rest of them are treated as low-resource datasets. We experiment our proposed methods on all low-resource datasets. 

We not only pre-train domain-based models on high-resource datasets, but also on all datasets except for Disease, Fiction, Tourism and Patent datasets because these datasets contain too few training data. For example, when experimenting on Social Media dataset, we use Social Media, MSRA, PKU and CTB8 to pre-train four domain-based models, respectively. The Social Media dataset provides in-domain knowledge and the rest of them provide out-of-domain knowledge.

\begin{table*}[!hbt]
\caption{Experimental results on six low-resource datasets. The proposed methods are described in Section 2. *We refer to PKU as the News dataset.   
}
\centering
\begin{tabular}{c|c|c|cccccc}
\hline
\multicolumn{1}{c|}{\multirow{1}{*}{}}&\multicolumn{1}{c|}{\multirow{1}{*}{Models}}&&\multicolumn{1}{c}{\multirow{1}{*}{Social Media}}&\multicolumn{1}{c}{\multirow{1}{*}{News}}&\multicolumn{1}{c}{\multirow{1}{*}{Tourism}}&\multicolumn{1}{c}{\multirow{1}{*}{Disease}}&\multicolumn{1}{c}{\multirow{1}{*}{Fiction}}&\multicolumn{1}{c}{\multirow{1}{*}{Patent}}\\

\hline
\multicolumn{1}{c|}{\multirow{12}{*}{\textbf{Baselines}}}&\multicolumn{1}{c|}{\multirow{3}{*}{Benchmark}}
&P&92.42&94.64&90.77&89.61&85.92&87.86\\
&&R&93.04&93.95&91.21&89.54&87.51&88.88\\
&&F&92.73&94.30&90.99&89.57&86.71&88.35\\
\cline{2-9}

&\multicolumn{1}{c|}{\multirow{3}{*}{Mixing}}
&P&92.21&94.25&91.33&88.81&90.92&90.21\\
&&R&93.18&93.66&91.80&90.15&89.88&90.64\\
&&F&92.69&93.96&91.57&89.47&90.40&90.43\\
\cline{2-9}
&\multicolumn{1}{c|}{\multirow{3}{*}{Bagging}}
&P&82.69&92.42&90.89&87.10&87.81&84.43\\
&&R&90.49&91.95&93.71&88.28&89.93&90.23\\
&&F&86.41&92.18&92.28&87.68&88.85&87.23\\
\cline{2-9}
&\multicolumn{1}{c|}{\multirow{3}{*}{Parameter Sharing}}
&P&92.94&95.55&94.81&89.23&89.84&90.56\\
&\multicolumn{1}{c|}{\multirow{3}{*}{~\cite{ChenSQH17}}}&R&93.42&95.20&95.45&90.30&91.67&91.77\\
&&F&93.18&95.38&95.13&89.76&90.80&91.16\\
\hline
\hline

\multicolumn{1}{c|}{\multirow{12}{*}{\textbf{Our Methods}}}&

\multicolumn{1}{c|}{\multirow{3}{*}{Gaussian-SNs}}
&P&92.03&95.14&87.67&87.10&85.21&86.40\\
&&R&92.89&94.73&89.66&88.28&87.15&88.05\\
&&F&92.46&94.93&88.66&87.68&86.17&87.22\\
\cline{2-9}

&\multicolumn{1}{c|}{\multirow{3}{*}{Concatenate-SNs}}
&P&92.38&96.03&95.37&91.55&94.71&92.05\\
&&R&93.84&95.64&95.97&92.43&95.39&93.80\\
&&F&93.10&95.83&95.67&91.99&95.05&92.91\\
\cline{2-9}

&\multicolumn{1}{c|}{\multirow{3}{*}{Tree-SNs}}
&P&92.74&96.12&95.46&91.28&\textbf{94.95}&\textbf{92.32}\\
&&R&94.26&95.61&96.24&92.13&\textbf{95.63}&\textbf{93.50}\\
&&F&93.50&95.87&95.86&91.70&\textbf{95.29}&\textbf{92.91}\\
\cline{2-9}

&\multicolumn{1}{c|}{\multirow{3}{*}{Sequence-SNs}}
&P&\textbf{93.13}&\textbf{96.16}&\textbf{95.61}&\textbf{91.69}&94.85&\textbf{92.03}\\
&&R&\textbf{94.14}&\textbf{95.64}&\textbf{96.20}&\textbf{92.50}&95.57&\textbf{93.82}\\
&&F&\textbf{93.63}&\textbf{95.90}&\textbf{95.91}&\textbf{92.09}&95.21&\textbf{92.91}\\
\cline{2-9}

\hline
\end{tabular}

\label{daimprovements}
\end{table*}

\subsection{Settings}

All datasets are preprocessed by replacing numbers and continuous English characters with special flags. All results are evaluated by $F_{1}$-score.

For training, we use mini-batch stochastic gradient
descent to minimize negative log-likelihood. Training is performed with shuffled mini-batches of size 15. Following previous works and experimental results on development datasets, both the character embedding dimension and the hidden dimension are set to be 100. All weight matrices, except for the bias vectors and word embeddings, are diagonal matrices and randomly initialized by normal distribution.

Experiments are performed on a commodity 64-bit Dell Precision T5810 workstation with one 3.0 GHz 16-core CPU and 64GB RAM. The C\# multiprocessing module is used in the experiment.

\begin{figure}[!hbt]
  \centerline{\includegraphics[width=0.95\linewidth]{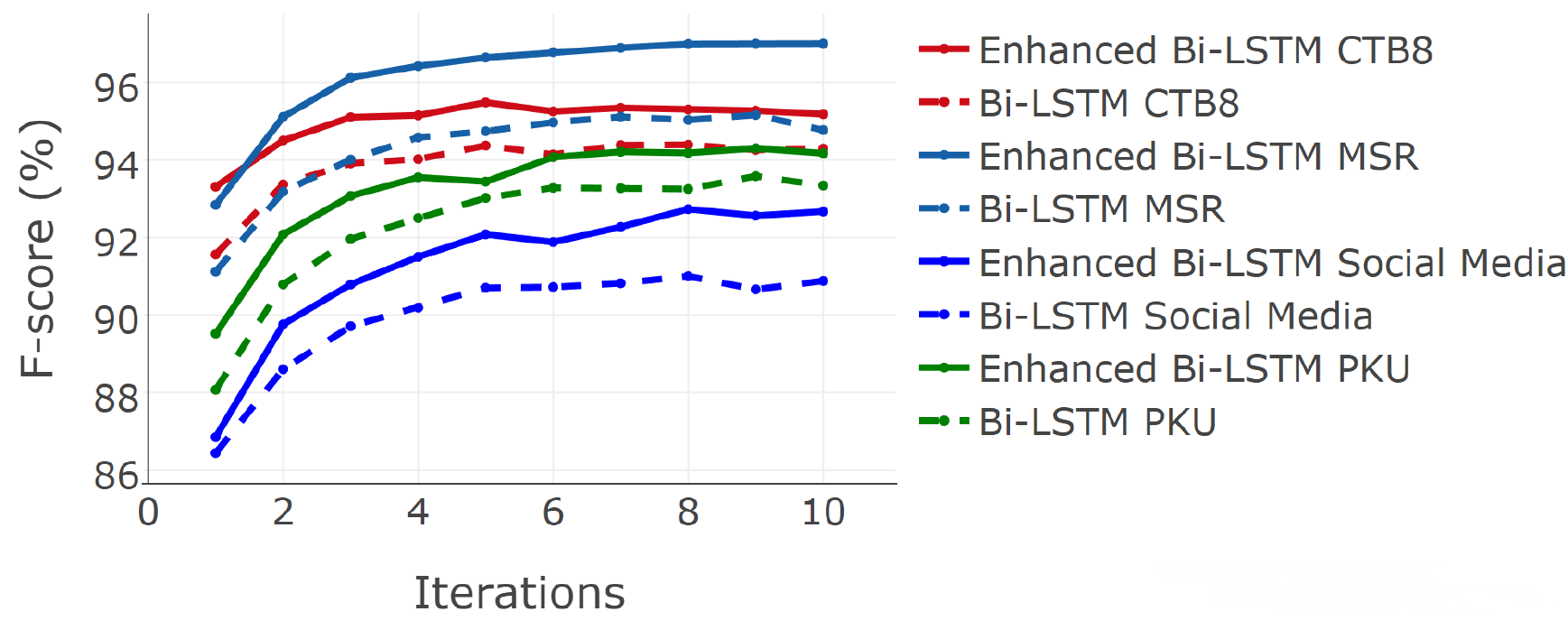}}
\caption{Comparisons between Bi-LSTM and Enhanced Bi-LSTM on CTB8, MSR, PKU and Social Media datasets.  }
\label{lstmandgrnn}
\end{figure}

\begin{table}
\caption{Results of Sequence-SNs with different orders of domain-based models on three datasets.
}
\centering
\begin{tabular}{c|c|c|c|c}

\hline
\multicolumn{2}{c|}{\multirow{1}{*}{Models}}&Social Media&News&Tourism\\

\hline

\multicolumn{1}{c|}{\multirow{3}{*}{Order-I}}&{P}&93.13&96.16&95.61\\
&R&94.14&95.64&96.20\\
&F&93.63&95.90&95.91\\
\hline
\multicolumn{1}{c|}{\multirow{3}{*}{Order-II}}&{P}&92.68&96.09&95.48\\
&R&94.26&95.51&96.15\\
&F&93.47&95.80&95.82\\
\hline

\multicolumn{1}{c|}{\multirow{3}{*}{Order-III}}&{P}
&92.86&96.27&95.58\\
&R&94.18&95.60&96.19\\
&F&93.52&95.94&95.89\\

\hline

\end{tabular}

\label{orders}
\end{table}

\subsection{Baselines}

In this paper, we report four baselines to compare with our proposed methods.

\begin{itemize}
\item \textbf{Benchmark} This model is trained on the target dataset. All out-of-domain datasets are not involved into the training. 

\item \textbf{Mixing} The model is trained on 
all available corpora, including in-domain and out-of-domain datasets. All datasets are involved into training directly. For this method, the shift of data distributions and the conflict of annotation guidelines are not fixed.

\item \textbf{Bagging} Bagging is a common method in ensemble learning. We integrate these domain-based models in a voting way. In our experiment settings, all domain-based models have the same voting weight.

\item \textbf{Parameter Sharing} We also include a common multi-task learning method to our baselines, Parameter Sharing~\cite{ChenSQH17}. It is the most commonly used approach in neural networks~\cite{duong,YangH16}. It is applied by sharing the parameters in the hidden layers among all tasks, while keeping several task-specific output layers.

\end{itemize}

\subsection{Results and Discussions}

Domain-based models can be realised with any sequence labelling structure. In this paper, we experiment two well performing models for CWS, Bi-LSTM and Enhanced Bi-LSTM. Bi-LSTM is a common sequence labelling model. Enhanced Bi-LSTM is the combination of Bi-LSTM and GRNN~\cite{chen} which is able to capture complex features.  The results are shown in Figure \ref{lstmandgrnn}. It can be clearly seen that Enhanced Bi-LSTM achieves the better results in all datasets. Therefore, we use Enhanced Bi-LSTM as our domain-based model in experiments.

The main results are shown in Table \ref{daimprovements}. It can be clearly seen that our proposed methods achieve large improvements over all baselines, which proves the effectiveness of our proposed methods.  Based on the results, some important findings are concluded as follows.

\textbf {``Dataset Bias'' and model conflicts decrease the performance.} 
Even with the use of multi-domain datasets, \textsl{Mixing }method does not beat \textsl{benchmark} which only uses the target dataset. The lower results reflect the negative influence of ``Dataset Bias''.  Moreover, \textsl{Bagging} performs poorly on most of datasets. The lower results indicate the problem of model conflicts.

\textbf {Parametric models achieve the better results than non-parametric models. } Gaussian-SNs achieve the worst performance among our proposed networks. The results are not surprising because this model mainly depends on $\sigma$ which is shared by all examples and tuned manually, rather than learned from the target dataset.

\textbf {Sequence-SNs achieve the best performance on most of datasets.} The best performance on most of datasets indicates the effectiveness of Sequence-SNs. Furthermore, shuffling the order of domain-based models does not bring too much result variance (shown in Table \ref{orders}). The maximum deviation is less than $0.2\%$. 


\textbf {Tree-SNs are still a good choice when a target dataset extremely lacks training data. } We find that Tree-SNs achieve the best performance on the two smallest datasets. It is mainly attributed to the model structure. Recursive cells in the same layer share the same parameters which would be updated many times in a training case. This mechanism is very useful in tasks with extremely insufficient data. Since the recursive cells treat the two inputs equally, it can not give preferential treatment to well-performing models during training, like Sequence-SNs do. It probably explains why sequence-SNs achieve the better results than Tree-SNs on the larger datasets.  

\begin{figure}
  \centerline{\includegraphics[width=0.90\linewidth]{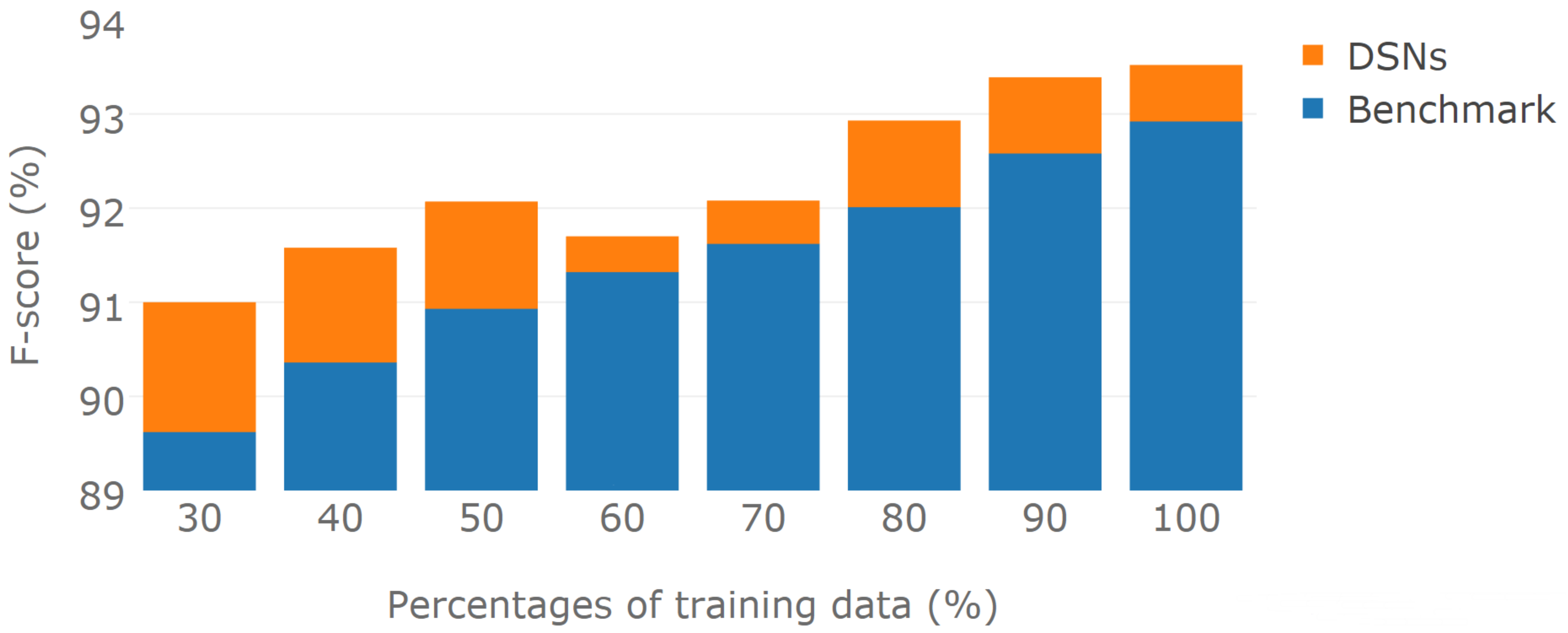}}
\caption{Results of deep stacking networks (Sequence-SNs) and \textsl{Benchmark} with different percentages of training data (Social Media). The orange bar represents the gap between  our proposed methods and \textsl{benchmark}.}
\label{percentage}
\end{figure}

\textbf{Our method is more effective in tasks with less labelled training data.} We also run our methods with different percentages of training data. Figure \ref{percentage} shows that the less training data there is, the higher performance improvements our methods gain. In particular, we achieve 1.4\% F-score improvement with 30\% labelled training data.

\begin{table}[!hbt]
\caption{Results of Sequence-SNs on the Social Media dataset with different domain-based models.
}
\centering
\begin{tabular}{|c|c|c|c|}
\hline
\multicolumn{1}{|c|}{\multirow{2}{*}{Domain-based Models}}&\multicolumn{3}{|c|}{\multirow{1}{*}{Social Media}}\\
\cline{2-4}
&P&R&F\\


\hline

MSR&92.36&93.55&92.95\\

PKU&92.81&93.37&93.09\\

CTB8&92.33&93.48&92.90\\

Social Media&92.41&93.04&92.73\\
\hline
MSR+Social Media&93.16&93.94&93.55\\

PKU+Social Media&92.97&93.98&93.47\\

CTB+Social Media&92.68&93.83&93.25\\
\hline
MSR+PKU+CTB8&92.97&94.25&93.51\\
\hline
All datasets&93.13&94.14&\textbf{93.63}\\
\hline

\end{tabular}

\label{diversedata}
\end{table}

\begin{table}[!hbt]
\caption{Comparisons with previous works on the PKU dataset.}
\centering
\begin{tabular}{|c|c|c|c|}
\hline
\multicolumn{1}{|c|}{\multirow{2}{*}{Models}}&\multicolumn{3}{|c|}{\multirow{1}{*}{PKU}}\\
\cline{2-4}
&P&R&F\\

\hline
Tseng et al. \cite{Tseng05}&*&*&95.0\\
Zhang et al. \cite{Zhang2006}&*&*&95.1\\
Zhang and Clark ~\cite{zhangclark2007}&*&*&94.5\\
Sun et al. \cite{SunWL12}&*&*&95.4\\
Pei et al. \cite{pei}&*&*&95.2\\
Ma and Hinrichs \cite{ma}&*&*&95.1\\
Zhang et al. \cite{Zhang2016Transition}&*&*&95.7\\
\textbf{Our Work}&95.6&96.2&\textbf{95.9}\\
\hline

\end{tabular}

\label{PKU}
\end{table}

%
%
%

\textbf{Our proposed methods have the strong conflict management skills. }We also explore different domain combinations, as shown in Table~\ref{diversedata}.  For example, ``MSR+PKU+CTB8'' represents that we integrate three models which are pre-trained on MSR, PKU and CTB8 datasets to improve the performance on the Social Media dataset. It can be seen that with the increase of the number of datasets, the performance becomes better. It shows that our proposed methods can learn knowledge from partially conflicting datasets.

Considering that the PKU dataset is used in previous works, we also compare our proposed methods with state-of-the-art methods. Table \ref{PKU}  shows that our proposed methods outperform previous works.   

\subsection{Case Study}
\begin{figure}
  \centerline{\includegraphics[width=0.9\linewidth]{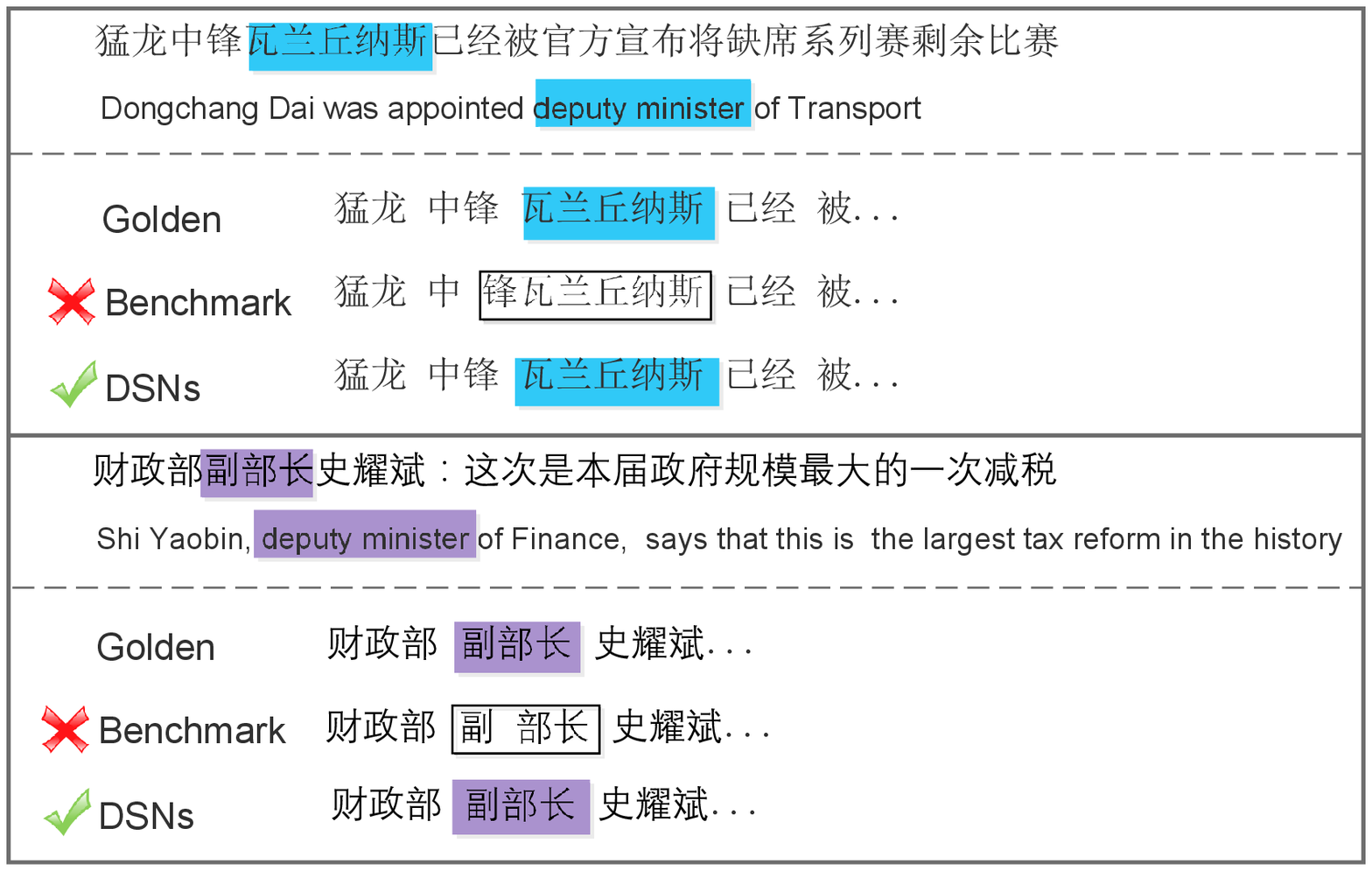}}
  \caption{Examples from the testing set of Social Media.  Words (shown in blue) are unseen in the labelled target data. The green check mark means a correct output and red cross means an incorrect output. In the top example, words shown in blue only appear in the out-of-domain datasets. In the bottom example, ``deputy minister'' has different annotation guidelines.  }
\label{casestudy}
\end{figure}

we consider several examples shown in Figure \ref{casestudy}. In the first example, ``Jonas Valanciunas'' (shown in blue) is unseen in the target dataset and only appear in out-of-domain corpora.  The right segmentation proves that our methods have the ability  to transfer knowledge from out-of-domain datasets. In the second example, ``deputy minister'' has different annotation guidelines as shown in Figure 1. The right segmentation shows the ability to handle the conflicts of annotation guidelines.  

\section{Related work}

Our work focus on domain adaptation and semi-supervised learning for neural word segmentation in Chinese social media. 

\textbf{Neural Word Segmentation.} 
A number of recent works attempted to extract features automatically by using neural networks~\cite{ZhengCX13,pei,ma,chen,Zhang2016Transition,Xu2016Dependency}. Pei et al.~\cite{pei} used CNN to capture local information within a fixed size window and proposed a tensor framework to capture the information of previous tags. Ma and Hinrichs~\cite{ma} proposed an embedding matching approach to CWS, which took advantage of distributed representations. The training and prediction algorithms had linear-time complexity. Chen et al.~\cite{chen} proposed gated recursive neural networks to model feature combinations of context characters. This gating mechanism was used in Cai and Zhao~\cite{Cai2016Neural} work. Zhang et al.~\cite{Zhang2016Transition} proposed a neural model for word-based Chinese word segmentation, rather than traditional character-based CWS, by replacing the manually designed discrete features with neural features
in a word-based segmentation framework.

\textbf{Transfer Learning in CWS.} 

\nocite{DBLP:journals/corr/XuS17}

Transfer learning aims to learn knowledge from different source domains to improve prediction results in a target domain. Domain adaptation has been successfully applied to many fields, such as machine learning~\cite{pan2011domain}, text sentiment classification~\cite{cWangM11}, image classification~\cite{DuanICML2012}, human activity classification~\cite{corr} and so on. Several methods have been proposed for solving domain adaptation problem in CWS. Peng et al.~\cite{Peng} presented the framework that easily supported the integration of domain knowledge in the form of multiple lexicons of characters and words. Jiang, Huang and Liu~\cite{JiangHL09} presented a simple yet effective strategy that transferred knowledge from a differently annotated corpus to the corpus with desired annotation. Zhang et al.~\cite{zhang2014type} used domain specific tag dictionaries and only unlabelled target domain data to improve target-domain accuracies. However, these domain adaptation methods involve in too many handcrafted features.

\section{Conclusions}

In this paper, we propose a novel stacking framework to improve low-resource CWS by incorporating multi-domain datasets. To reduce model conflicts, we design various stacking networks, including Gaussian-SNs, Concatenate-SNs, Sequence-SNs and Tree-SNs. Experimental results show some interesting and important findings. First, our proposed methods achieve significant performance gains on all datasets compared with the strong baselines. Second, parametric models achieve the better results than non-parametric models (Gaussian-SNs). Third, Sequence-SNs show the best performance on most of datasets and Tree-SNs are more suitable for datasets which extremely lack training data.

\IEEEpeerreviewmaketitle


\bibliographystyle{IEEEtran}
\bibliography{ieeeexample}

\end{document}